\documentclass[letterpaper, 10 pt, conference]{ieeeconf}  

\IEEEoverridecommandlockouts                              

\usepackage{cite}
\usepackage{hyperref}
\usepackage{graphics} 
\usepackage{epsfig} 
\usepackage{amsmath} 
\usepackage{amssymb}  
\usepackage{booktabs}
\usepackage{multirow} 
\usepackage{xcolor} 
\usepackage{tabularx}
\usepackage{threeparttable}
\usepackage{bm}
\usepackage{verbatim}
\usepackage{makecell}
\usepackage{cleveref}

\newcolumntype{Y}{>{\centering\arraybackslash}X}

\title{\LARGE \bf
AERO-MPPI: Anchor-Guided Ensemble Trajectory Optimization for Agile Mapless Drone Navigation
}

\author{Xin Chen, Rui Huang, Longbin Tang, Lin Zhao
\thanks{The authors are with the Department of Electrical and Computer Engineering, National University of Singapore, Singapore 117583. Email: 
        \{e1499417, ruihuang, longbin\}@u.nus.edu, zhaolin@nus.edu.sg. This work was supported by the Singapore Ministry of Education Tier 2 Academic Research Fund (T2EP20123-0037, T2EP20224-0035). Corresponding author: Lin Zhao.}
}

\begin{document}

\maketitle
\thispagestyle{empty}
\pagestyle{empty}

\begin{abstract}
Agile mapless navigation in cluttered 3D environments poses significant challenges for autonomous drones. Conventional mapping–planning–control pipelines incur high computational cost and propagate estimation errors. We present AERO-MPPI, a fully GPU-accelerated framework that unifies perception and planning through an anchor-guided ensemble of Model Predictive Path Integral (MPPI) optimizers. Specifically, we design a multi-resolution LiDAR point-cloud representation that rapidly extracts spatially distributed “anchors” as look-ahead intermediate endpoints, from which we construct polynomial trajectory guides to explore distinct homotopy path classes. At each planning step, we run multiple MPPI instances in parallel and evaluate them with a two-stage multi-objective cost that balances collision avoidance and goal reaching. Implemented entirely with NVIDIA Warp GPU kernels, AERO-MPPI achieves real-time onboard operation and mitigates the local-minima failures of single-MPPI approaches. Extensive simulations in forests, verticals, and inclines  demonstrate sustained reliable flight above 7 m/s, with success rates above 80\% and smoother trajectories compared to state-of-the-art baselines. Real-world experiments on a LiDAR-equipped quadrotor with NVIDIA Jetson Orin NX 16G confirm that AERO-MPPI runs in real time onboard and consistently achieves safe, agile, and robust flight in complex cluttered environments. Code is available at \url{https://github.com/XinChen-stars/AERO_MPPI}.
\end{abstract}


\section{Introduction}

Agile flight in cluttered, unknown environments remains a significant challenge for autonomous aerial robots. Conventional autonomous flight stacks~\cite{fastplanner2019robust,topoplanner2020,egoplanner2021,liu2024integrated} typically employ a sequence of modules, such as mapping, safe-flight corridor generation, planning, control, and so on. These pipelines perform well for low-speed flight in sparse obstacle fields. However, such designs typically demand heavy onboard computation and introduce cumulative latency that limits flight speed. Moreover, the sequential processing propagates estimation and planning errors, making the system prone to crashes in densely cluttered environments. 

\begin{figure}[thpb]
    \centering
    \includegraphics[width=1\linewidth]{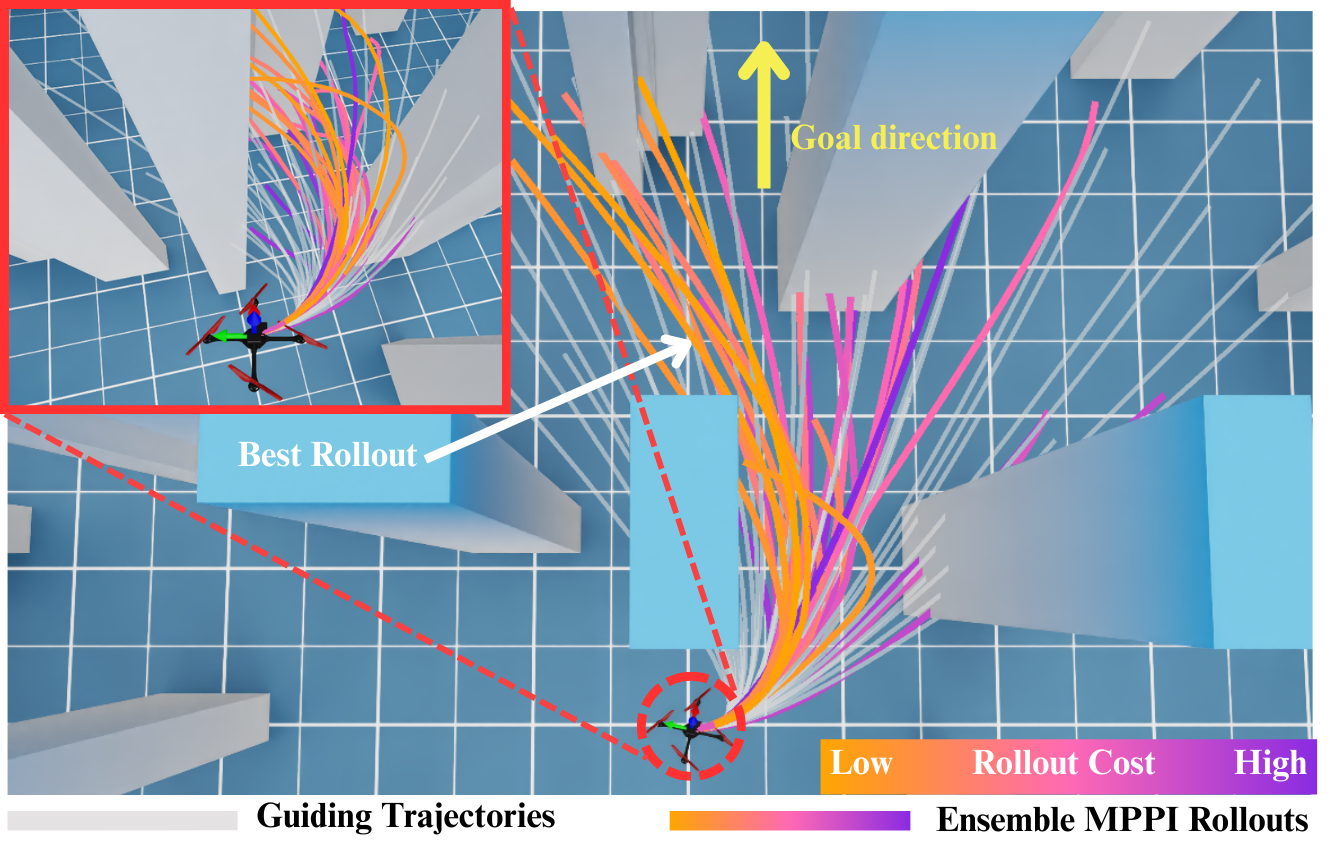} 
    \vspace{-8 mm}
    \caption{AERO-MPPI navigating a challenging obstacle environment. The algorithm samples spread anchor points and generates corresponding guiding trajectories (gray lines) on the LiDAR point cloud. These trajectories are then optimized by parallel MPPI instances to produce multiple trajectory proposals (\textcolor{orange}{orange}–\textcolor{purple}{purple} curves). The best is the one with the lowest cost.}
    \label{fig:avoidance}
    \vspace{-7 mm}
\end{figure}

Mapless navigation generates and executes collision-free motions using only local, onboard perception (e.g., LiDAR point clouds). This avoids the error accumulation inherent in maintaining a global map and thereby enhances flight agility. Among sampling-based trajectory optimizers, Model Predictive Path Integral (MPPI) control~\cite{mppi2018} is a strong candidate because it flexibly accommodates diverse objectives, system dynamics, and constraints. More importantly, its sampling-based optimization is well-suited to parallelization.

\begin{figure*}[t]
    \centering
    \includegraphics[width=0.95\textwidth]{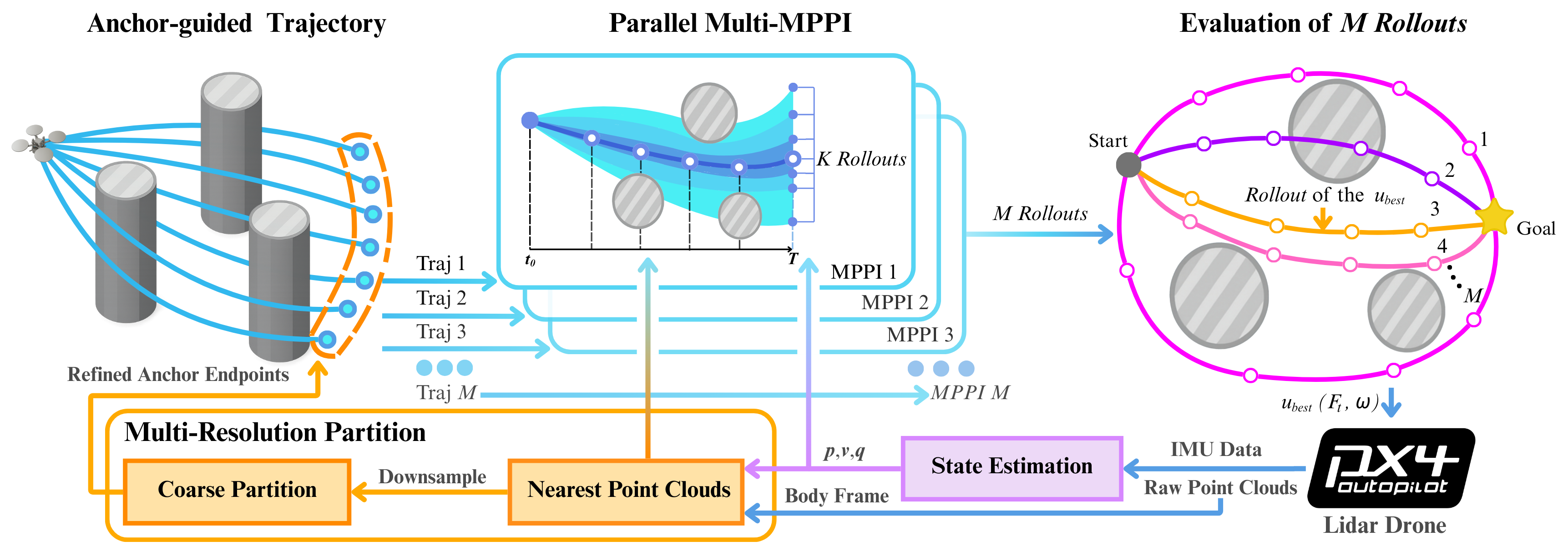}
    \vspace{-4 mm}
    \caption{\textbf{AERO-MPPI System Overview.}  We perform state estimation by fusing IMU data with raw LiDAR point clouds to recover the system state $\left(\mathbf{p}, \mathbf{v}, \mathbf{q}\right)$. To enhance efficiency, we design a multi-resolution partitioning scheme that downsamples the nearest point clouds onto coarsely defined partitions to obtain refined anchor endpoints. For free-space exploration, the drone leverages the refined endpoints together with the current state to compute guiding trajectories, which serve as initial guesses for parallel multi-MPPI optimization. After obtaining the candidate control sequences from multi-MPPI, we evaluate them through explicit rollouts and select the best control input.
} 
    \label{fig:overview}
    \vspace{-6 mm}
\end{figure*}

However, MPPI-based planning directly on raw point clouds presents two major challenges. First, the repeated evaluation of collision costs against dense point sets is computationally intensive and can easily break real-time operation. Second, a single-start MPPI is prone to poor exploration in cluttered environments: with only one goal point guidance, MPPI rollouts often remain confined to a single homotopy class, making the planner sensitive to cost shaping and initial seeds and thus prone to local traps. Increasing the sampling covariance can partly alleviate this issue, but it typically slows the optimization process and can even lead to divergence or instability.

To address these challenges, we propose AERO-MPPI, a parallel ensemble of MPPI optimizers guided by diversified look-ahead references we call \textit{anchors} (see the overall diagram in Fig.~\ref{fig:overview}). An anchor is a per-cycle, goal-directed look-ahead point from which we generate a guiding trajectory (grey curves in Fig.~\ref{fig:avoidance}); Anchors are sampled along the goal direction with a configurable look-ahead distance and angular spread, ensuring that the resulting generated guiding trajectories span distinct corridors around obstacles and enable more robust navigation. In addition, we build a multi-resolution LiDAR representation that supports fast generation of these anchor-guided trajectories. They serve as high-quality initial guesses for subsequent MPPI optimizations. At each planning step, AERO-MPPI samples anchors, generates the corresponding guiding trajectories, runs the MPPI instances concurrently, and selects the best proposal under a multi-objective cost that balances collision avoidance and goal progress. The entire planning pipeline is implemented with NVIDIA Warp~\cite{warp2022} GPU kernels, fully leveraging large-scale parallelism for onboard, real-time performance.

Our main contributions are summarized as follows:

\begin{itemize}
    \item We propose AERO-MPPI, an anchor-guided ensemble of MPPI optimizers for parallel trajectory optimization. This approach mitigates the local-minima issues common in traditional MPPI and significantly improves navigation robustness during high-speed flight.
    \item We implement the full perception-to-planning pipeline with GPU kernels (NVIDIA Warp), enabling real-time, high-speed flight with onboard computation.
    \item We conducted extensive validation in simulation across challenging scenarios. Results show our method achieves improved agility, success rate, and smoothness compared with baseline planners~\cite{egoplanner2021,topoplanner2020,fastplanner2019robust,liu2024integrated}, while maintaining real-time performance, safety, and deployability.

\end{itemize}

\section{RELATED WORK}
Popular drone navigation algorithms mostly relies on various forms of local maps for collision avoidance trajectory planning. For example, Fast-Planner~\cite{fastplanner2019robust} performs unconstrained trajectory optimization on an online-updated Euclidean Signed Distance Field (ESDF) for collision avoidance. Following Fast-Planner, Topo-Planner~\cite{topoplanner2020} generates and selects among multiple homotopy-aware trajectories on the ESDF, improving global optimality in cluttered environments. Furthermore, Ego-Planner~\cite{egoplanner2021} performs local greedy trajectory optimization directly on the occupancy map without explicit ESDF construction. In addition, Ego-Planner v2 further improves trajectory quality and generation efficiency by adopting MINCO~\cite{minco2022} as the trajectory parameterization. Moreover, IPC~\cite{liu2024integrated} constructs a local map and a Safe Flight Corridor (SFC)~\cite{sfc2017}, and enforces SFC as hard constraints in a linear model predictive control (MPC) framework for integrated planning and control. These methods still require frequent ESDF updates or SFC construction, which require significant computation and induce large latency, thus limiting the flight speed in cluttered scenarios.
 
The aforementioned works rely on gradient information from explicit maps. In contrast, MPPI control~\cite{mppi2018} is a sampling-based, model-predictive method that repeatedly draws random control sequences, simulates (rolls out) the system’s nonlinear dynamics under those controls, and evaluates a cost for each trajectory. The controls are then updated via a path-integral weighted average of the samples, favoring lower cost trajectories and yielding an optimized control action at each planning step. This formulation naturally handles nonlinear dynamics and non-smooth obstacle costs, while GPU parallelization enables thousands of rollouts in real time.

Recent works have implemented GPU-accelerated MPPI for onboard quadrotor computation, enabling real-time collision avoidance~\cite{mppiimple2024}. To improve sampling efficiency and trajectory smoothness, spline-interpolated MPPI with Stein variational inference has been proposed~\cite{splinemppi}. Other variants, such as Tube-MPPI~\cite{tubemppi2022}, Robust-MPPI (RMPPI)~\cite{rmppi2023}, and Risk-Aware MPPI (RA-MPPI)~\cite{ramppi2023}, enhance robustness to disturbances and model uncertainties. To provide stronger safety guarantees, Control Barrier Functions (CBFs)~\cite{cbfmppi2022} have been incorporated into MPPI in methods like Shield-MPPI~\cite{shieldmppi2023} and SCBF-MPPI~\cite{scbfmppi2023} to filter unsafe trajectory samples. 

However, all these approaches rely on a single MPPI instance, which often leads to local minima and conservative behavior. They also depend on explicit mapping modules to evaluate risk and safety, which are expensive to construct in unknown or dynamic environments. In contrast, our AERO-MPPI fosters global exploration and can plan more diverse homotopy trajectories, enabling more effective obstacle-avoidance navigation.

\section{Preliminaries}
\subsection{Quadrotor Model}
The following quadrotor model is employed:
\begin{equation}
\dot{\boldsymbol{p}}=\boldsymbol{v},\;
\dot{\boldsymbol{q}}=\tfrac{1}{2}\,\boldsymbol{q}\odot
\begin{bmatrix}0\\\boldsymbol{\omega}\end{bmatrix},\;
\dot{\boldsymbol{v}}=\tfrac{1}{m}\,\boldsymbol{R}(\boldsymbol{q})
\begin{bmatrix}0\\0\\F_t\end{bmatrix}+\boldsymbol{g},
\label{eq:dynamics}
\end{equation}
where $\boldsymbol{p}, \boldsymbol{v} \in \mathbb{R}^3$ denote the position and velocity of the body frame $B$ with respect to the world frame $W$, respectively. The body’s orientation relative to an inertial/world frame is represented by the unit quaternion $\boldsymbol{q}$ with $|\boldsymbol{q}| = 1$, and $\boldsymbol{\omega}$ is the angular velocity of the body expressed in the body frame, $\boldsymbol{R}(\boldsymbol{q})$ is the rotation matrix corresponding to $\boldsymbol{q}$, $\odot$ denotes quaternion multiplication, $m$ is the drone mass, and $\boldsymbol{g}$ is the gravitational acceleration vector. We adopt CTBR (collective thrust and body rate) control with input $\boldsymbol{u} := (F_t, \boldsymbol{\omega}) \in \mathbb{R}^4$. For brevity, we denote $\boldsymbol{x} := [\boldsymbol{p}, \boldsymbol{q}, \boldsymbol{v}]$.

\subsection{Model Predictive Path Integral Control}

MPPI is an optimal control framework that solves nonlinear and nonconvex problems through Monte Carlo sampling rather than gradient-based optimization. This sampling-based formulation allows direct incorporation of complex dynamics, high-dimensional constraints, and cost functions with sparse or nonexistent gradients. With receding-horizon execution, MPPI continuously replans using updated perception information, making it well suited for motion planning in cluttered environments. 

At each control step, MPPI predicts $K$ candidate trajectories (rollouts) over $N$ discrete time steps (prediction horizon). The current state is $\boldsymbol{x_0}$, and the nominal control sequence is $\boldsymbol{u}^{\mathrm{nom}} = (\boldsymbol{u}_0^{\mathrm{nom}}, \dots, \boldsymbol{u}_{N-1}^{\mathrm{nom}})$. Gaussian disturbances $\boldsymbol{\delta}_j^k \sim \mathcal{N}(\mathbf{0}, \Sigma)$ 
are added to the nominal controls to generate perturbed controls 
$\boldsymbol{u}_j^k = \boldsymbol{u}_j^{\mathrm{nom}} + \boldsymbol{\delta}_j^k$ for rollout $k$,
where $\Sigma \in \mathbb{R}^{n_u \times n_u}$ is the covariance matrix.

Each rollout $k$ is simulated forward using the continuous-time dynamics \eqref{eq:dynamics}, which are discretized with a time step $\Delta t$ using the fourth-order Runge--Kutta (RK4) integration method:
\begin{equation}
\boldsymbol{x}_{j+1}^k = \boldsymbol{x}_j^k + f_{\mathrm{RK4}}\!\left( \boldsymbol{x}_j^k, \boldsymbol{u}_j^k, \Delta t \right).
\label{eq:mppi_rollout_simplified}
\end{equation}
where $j\in\{0,\ldots,N-1\}$ denotes the time step along the prediction horizon.

For each trajectory, a task-specific cost $S_k$ is computed, and the $K$ costs are normalized to generate a distribution $(\varepsilon_1,\varepsilon_2,\ldots,\varepsilon_K)$:
\begin{equation}
\begin{aligned}
\rho &= \min\{S_1, \cdots, S_K\},\\
\eta &= \sum_{k=1}^K \exp\!\left( -\tfrac{1}{\lambda}(S_k - \rho) \right), \\[4pt]
\varepsilon_k &= \frac{\exp\!\left( -\tfrac{1}{\lambda} (S_k - \rho) \right)}{\eta},
\end{aligned}
\label{eq:mppi_weights}
\end{equation}
where $\lambda>0$ is the so-called temperature parameter that controls the spread of the distribution. It balances exploration and exploitation: as $\lambda \to 0$, the weights collapse to the minimum-cost trajectory, 
while larger $\lambda$ values distribute weights more evenly across rollouts.

The nominal controls are updated by the weighted sum of the sampled disturbances $\boldsymbol{\delta} \boldsymbol{u}^k = (\boldsymbol{\delta}_{0}^k,\ldots,\boldsymbol{\delta}_{N-1}^k)$:
\begin{equation}
\boldsymbol{u}^{\mathrm{nom}} \gets \boldsymbol{u}^{\mathrm{nom}} + \sum_{k=1}^K \varepsilon_k \, \boldsymbol{\delta} \boldsymbol{u}^k.
\label{eq:mppi_update_simplified}
\end{equation}

This update shifts $\boldsymbol{u}^{\mathrm{nom}}$ toward the controls of lower-cost rollouts, iteratively improving the planned trajectory until convergence.

\begin{figure}[t]
    \centering
    \includegraphics[width=1.0\linewidth]{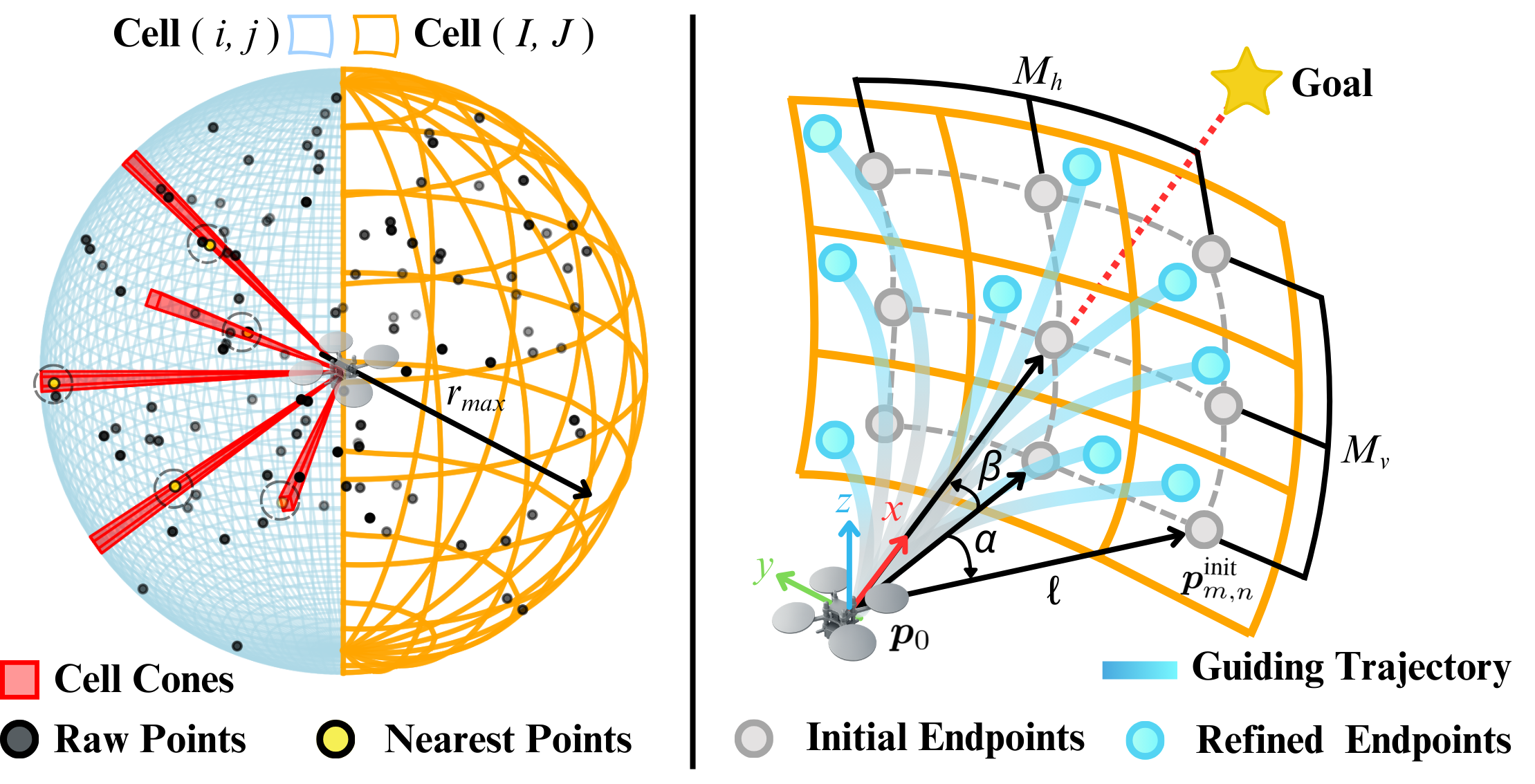}
    \vspace{-6 mm}
    \caption{\textbf{Multi-resolution partition for generating guiding trajectories.} Black points denote raw point clouds within the sensing sphere. The left hemisphere illustrates fine-grained cells, where selected cells are highlighted by the corresponding red spherical cones extending to the nearest LiDAR points (yellow). The right hemisphere depicts a coarse-grained partition used to adjust the initial anchor endpoints (gray) to refined endpoints (blue). Facilitated by this multi-resolution partition, guiding trajectories are generated efficiently from the current drone's pose to the refined endpoints.}
    \label{fig:percepetion_guidance}
    \vspace{-6 mm}
\end{figure}
\begin{figure*}[th] 
    \centering
    \includegraphics[width=1.0\textwidth]{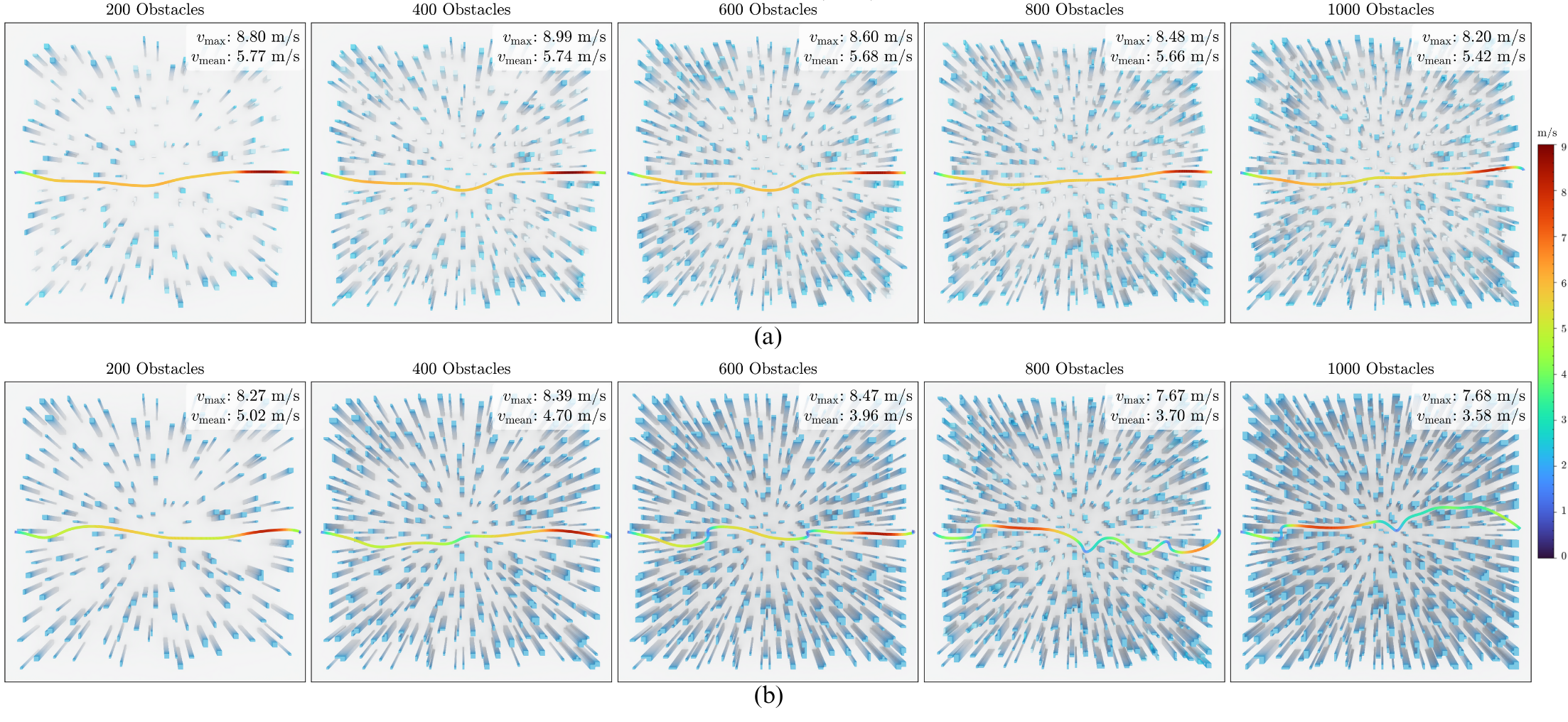}
    \vspace{-6 mm}
    \caption{  
\textbf{Comparison across obstacle numbers (200–1000) and height ranges (1–6\,m) in a \(40\,\mathrm{m} \times 40\,\mathrm{m}\) map.} 
(a) Randomized heights between 1–6 m: the five subfigures correspond to obstacle counts from 200 to 1000. 
With 1000 obstacles (each with a width ranging from 0.4 m to 1.1 m), the maximum velocity reaches 8.20 m/s and the mean velocity is 5.42 m/s, demonstrating the agility of the proposed AERO-MPPI. 
(b) Fixed height of 6 m: the uniform obstacle height prevents the drone from exploiting $z$-axis variation for avoidance. 
This configuration is more challenging, with narrow gaps that often cause local minima, reducing the maximum velocity to 7.68 m/s.
}   
\label{fig:different_obstacles}
\vspace{-6 mm}
\end{figure*}

\section{AERO-MPPI}

AERO-MPPI consists of three main components: 
1) Multi-resolution partitioning of the environment;
2) Anchor-guided trajectory generation; 
3) Parallel MPPI ensemble trajectory optimization. 
An overview of the pipeline is shown in Fig.~\ref{fig:overview}.

\subsection{Multi-Resolution Partition}
\label{sec:multi_resolution}
We build both a high-resolution partition and a coarse partition to facilitate anchor points generation and collision evaluation. The high-resolution partition decomposes the local search space into spherical cones, and the distance of the nearest obstacle in each cone is computed. The coarse partition is mainly used to generate anchor points to probe distinct corridors around obstacles. An illustration of the partition is illustrated in the hemispherical view of Fig.~\ref{fig:percepetion_guidance}.

\subsubsection{High-resolution partition (3$^\circ$)}
We adopt the lightweight Livox Mid-360 LiDAR, which features a non-repetitive scanning pattern with a 10\,Hz scan rate. Compounded with the drone’s fast motion, typically a single frame is insufficient to provide reliable coverage of the environment. Therefore, we accumulate the most recent 10 consecutive frames (200{,}000 points) and project them into the current body frame using the transforms from state estimation, yielding the accumulated point-cloud set $\mathcal{P}_t$. For onboard computation, we replan at 50 Hz using the latest drone pose provided by the PX4 EKF, which fuses IMU measurements with Fast-LIO localization~\cite{fastlio22022}. The perception range for processing the point-cloud is limited to $r_{\max}=10$\,m. Centered at the drone body frame, we discretize the spherical Field of View (FOV) into an angular grid at a $3^\circ$ resolution, resulting in $120{\times}60$ cells. Let $(\alpha_i,\beta_j)$ denote the azimuth and elevation of Cell $(i,j)$ with spherical cone $C_{i,j}, \; i\in\{1,\cdots,120\}, \; j\in\{1,\cdots,60\}$. For each cone, we compute the nearest distance to the obstacle
\begin{equation}
r_{i,j} \;:=\; \min_{\mathbf{p}_c\in \mathcal{P}_t \cap C_{i,j}} \|\mathbf{p}_c\|,
\end{equation}
where $\mathbf{p}_c$ denotes the coordinate of a point in the cloud expressed in the drone body frame.

The unit direction $\boldsymbol{d}_{i,j}\in\mathbb{R}^3$ of Cell $(i,j)$ is
\begin{equation}
\boldsymbol{d}_{i,j} \;:=\;
\begin{bmatrix}
\cos\beta_j \cos\alpha_i\\
\cos\beta_j \sin\alpha_i\\
\sin\beta_j
\end{bmatrix}.
\label{eq: unit direction}
\end{equation}

\subsubsection{Coarse partition  (18$^\circ$)}
We aggregate the \(3^\circ\)-resolution cells using \(6{\times}6\) pooling to form an \(18^\circ\)-resolution grid with a total of \(20{\times}10\) cells. Let \(\Omega_{I,J}\) denote the set of high-resolution cell indices pooled into the coarse cell \((I,J)\).

For each Cell \((I,J)\), we can compute the \textit{coarse safe point} $\boldsymbol{v}_{I,J}$ as
\begin{equation}
\begin{aligned}
(i^\star,j^\star) &= \arg\max_{(i,j)\in\Omega_{I,J}} r_{i,j}, \\[6pt]
\boldsymbol{v}_{I,J} &= r_{i^\star,j^\star}\,\boldsymbol{d}_{i^\star,j^\star},
\end{aligned}
\label{eq:safe-vector}
\end{equation}
where $(i^\star,j^\star)$ is the index with the maximum safe range in $\Omega_{I,J}$ and $\boldsymbol{d}_{i,j}$ is the unit direction vector of Cell $(i,j)$. Slightly abusing the notation, we denote the coarse safe direction of Cell\((I,J)\) by $\boldsymbol{d}_{I,J}:=\boldsymbol{d}_{i^\star,j^\star}$. 

\subsection{Anchor-Guided Trajectory Generation}
\label{sec:guiding_traj}
To broaden the search for distinct corridors around obstacles and reduce the risk of local minima, we first generate look-ahead anchor points. These anchors serve as endpoints for constructing polynomial trajectories. In~\Cref{sec:parallel_multi_mppi}, an ensemble of MPPI instances is then executed in parallel to track these trajectories.
\subsubsection{Sampling of the initial endpoints}
As illustrated in Fig.~\ref{fig:percepetion_guidance}, we set a look-ahead distance of \(\ell = 5\,\mathrm{m}\). At this distance, a grid of \(M_h \times M_v\) cells is first picked centered on the drone’s current heading direction from the coarse partition.  
We then rotate the entire grid so that its central axis aligns with the goal direction, defined by the azimuth angle \(\alpha\) and elevation angle \(\beta\) in the figure.  
Each gray dot in the diagram marks the center of a coarse cell and is used as an \emph{initial endpoint} for trajectory generation.  
We further refine these initial endpoints (highlighted by the cyan circles) according to the maximum safe distance and construct the \emph{guiding trajectories} (shown in blue) in the next subsection.

\begin{figure*}[t] 
    \centering
    \includegraphics[width=1.0\textwidth]{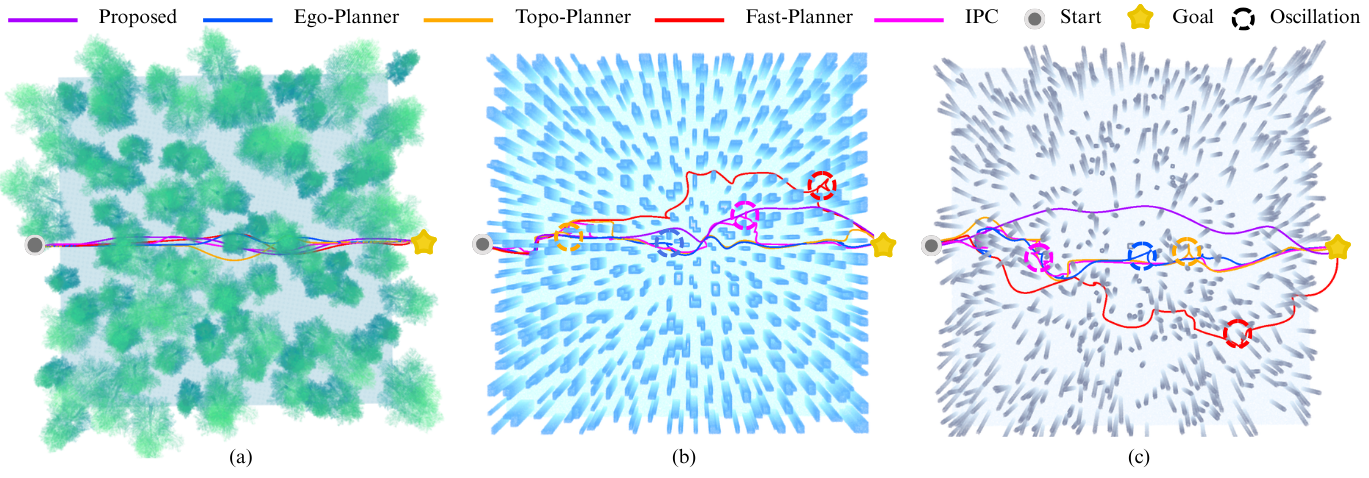}
    \vspace{-6 mm}
    \caption{   
    \textbf{Benchmark results with previous works \cite{egoplanner2021,topoplanner2020,liu2024integrated,fastplanner2019robust} under different scenarios in a \(40\,\mathrm{m} \times 40\,\mathrm{m}\) map.} 
    (a) Forest: All methods generate smooth trajectories of approximately the same length. 
    (b) Verticals: Mapping-based methods exhibit frequent replanning, leading to oscillatory and non-smooth trajectories, as tall vertical obstacles often cause severe occlusions. In contrast, our method directly leverages local perception for optimization, ensuring continuity and smoothness. 
    (c) Inclines: Similar to the Verticals case, frequent replanning occurs, and some trajectories even become trapped in local geometric pitfalls (e.g., small holes or narrow gaps), causing stagnation at local optima. Our method employs explicit parallel optimization and robust sampling to effectively avoid such traps.
    }
    \label{fig:simulated_comparison}
    \vspace{-6 mm}
\end{figure*}

\subsubsection{Refining the initial endpoints}
The initial endpoint is refined by re-aligning it to the direction of the maximum safe distance within its Cell \((I,J)\). Specifically, with the drone's position denoted by $\mathbf{p}_0$, $\forall I\in \{1,\ldots,M_h\}, J\in \{1,\ldots,M_v\}$, we have the refined endpoints calculated by
\begin{equation}
\boldsymbol{p}^{\text{ref}}_{I,J}
= \boldsymbol{p}_0 + \ell \cdot \boldsymbol{d}_{I,J},
\label{eq:anchor-point}
\end{equation}
where $\boldsymbol{d}_{I,J}$ is the unit direction vector as defined in~\eqref{eq:safe-vector}. 

This enforces that anchors strictly follow the locally safest heading provided by each cell in the coarse partition.

\subsubsection{Building Polynomial Guiding Trajectory}
Each guiding trajectory uses three fifth-order polynomials $f_\mu(t)$, $\mu\in\{x,y,z\}$, over horizon $T=N\Delta t$:
\begin{equation}
f_\mu(t) \;=\sum_{i=0}^{5}\; a_{i,\mu}t^i, \quad \forall t\in [0,T]
\label{eq:polynomial}
\end{equation}

We note that the coefficients $\{a_{i,\mu}\}_{i=0}^{5}$ in \eqref{eq:polynomial} are uniquely determined by the boundary conditions on position, velocity, and acceleration at the start and end of the horizon. This results in a closed-form polynomial solution that generates smooth, dynamically feasible guiding trajectories aligned with the coarse safe directions and distributed around the goal direction.

\subsection{Parallel MPPI Ensemble Optimization}
\label{sec:parallel_multi_mppi}
Unlike previous works~\cite{cbfmppi2022,shieldmppi2023, mppiimple2024} that rely on a single MPPI and often fall into local minima, our approach employs multiple MPPI instances running in parallel to enhance the exploration.  Specifically, we leverage the guiding trajectories as the tracking references to initialize $M:=M_h \times M_v$ MPPIs in parallel. 

Furthermore, prior methods are typically amenable to only a single-objective cost, which can make the planner either overly conservative or overly aggressive and ultimately lead to suboptimal performance. Our MPPI ensemble addresses this with a two-stage optimization: In Stage~I, each MPPI instance optimizes its guiding trajectory via sampling under a composite cost $S^{(1)}$, obtaining the optimized control sequence along the prediction horizon. In Stage~II, we roll out the $M$ optimized control sequences via RK4 integration of the drone dynamics, re-evaluate them with a simplified cost $S^{(2)}$, and select the optimal rollout control sequence for execution. This dual-stage design preserves directional diversity while enforcing an explicit safety filter. The two stage costs are defined respectively as follows:
\begin{equation}
\begin{aligned}
S^{(1)} &= 
J_{\mathrm{track}} + J_{\mathrm{vnorm}} + J_{\mathrm{ctrl}} 
+ S^{(2)}, \\[4pt]
S^{(2)} &= 
J_{\mathrm{goal}} + J_{\mathrm{col}}.
\end{aligned}
\label{eq:total cost}
\end{equation}

Each cost term in~\eqref{eq:total cost} is defined below. Matrices of the form $\boldsymbol{Q_*}$ denote the weighting matrices, with subscripts indicating the corresponding cost components.

\subsubsection{Guiding-Trajectory Tracking Cost}
To keep rollouts close to the chosen guide, we penalize the position deviation from the guiding trajectory:
\begin{equation}
J_{\mathrm{track}}
\;=\;
\sum_{t=0}^{N-1}\;
{\boldsymbol{Q}_{track}}\big\|\boldsymbol{p}_t - \boldsymbol{p}_{\mathrm{traj}}(t)\big\|,
\end{equation}
where $\mathbf{p}_t$ and $\mathbf{p}_{\mathrm{traj}}(t) = [f_x(t), f_y(t), f_z(t)]$ denote, respectively, the MPPI-predicted (rollout) position and the guiding-trajectory position at time step~$t$.

\subsubsection{Velocity Norm Cost}
To discourage excessively large velocities and ensure safe, we penalize the squared norm of the velocity along the rollouts:
\begin{equation}
J_{\mathrm{vnorm}}
\;=\;
\sum_{t=0}^{N-1}\;
{\boldsymbol{Q}_{vnorm}}\big\|\boldsymbol{v}_t\big\|^{2},
\end{equation}
where $\boldsymbol{v}_t$ is the MPPI predicted (rollout) velocity at time step~$t$.

\subsubsection{Control Cost}
To limit effort and smooth inputs, we penalize control magnitude and its rate:
\begin{equation}
J_{\mathrm{ctrl}}
\;=\;
\sum_{t=0}^{N-2}\;
{\boldsymbol{Q_c}}\big\|\boldsymbol{u}_t\big\|^{2}
\;+\;
\sum_{t=1}^{N-2}\;
{\boldsymbol{Q}_{c\Delta}}\big\|\Delta\boldsymbol{u}_t\big\|^{2},
\end{equation}
where $\Delta \boldsymbol{u}_t \triangleq \boldsymbol{u}_t - \boldsymbol{u}_{t-1}$, with $\boldsymbol{u}_{t-1}$ denoting the previously applied control.

\subsubsection{Goal Cost}
To bias the rollouts toward the goal, we penalize both running and terminal errors in position, velocity, and attitude:
\begin{equation}
\begin{aligned}
J_{\mathrm{goal}}
\;=\;
\sum_{t=0}^{N-1}
\Big(
&{\boldsymbol{Q}_{p}}\big\|\boldsymbol{p}_t - \boldsymbol{p}_{\mathrm{goal}}\big\|
\;+\;
{\boldsymbol{Q}_{v}}\big\|\boldsymbol{v}_t - \boldsymbol{v}_{\mathrm{goal}}\big\|
\\
&+\;
\boldsymbol{Q}_{q}\;
\big\|\,\boldsymbol{R}(\boldsymbol{q}_t)\,\boldsymbol{R}(\boldsymbol{q}_{\mathrm{goal}})^{\top} - \boldsymbol{I}\,\big\|_{F}
\Big),
\end{aligned}
\end{equation}
where $\boldsymbol{p}_{\mathrm{goal}}, \boldsymbol{v}_{\mathrm{goal}}, \boldsymbol{q}_{\mathrm{goal}}$ denote the goal position, velocity, and quaternion, respectively; $\boldsymbol{q}_t$ is the quaternion at time step~$t$; $\boldsymbol{I}$ is the identity matrix; and \(\|\cdot\|_{F}\) denotes the Frobenius norm.

\subsubsection{Collision Cost}
We penalize insufficient clearance by measuring the Euclidean distance from the rollout position $\boldsymbol{p}_t$ at time step~$t$ to the nearest observed obstacle point $\boldsymbol{p}_{\mathrm{obs}}$ in the high-resolution ($3^\circ$) partition. We define the nearest distance as \(d_t = \min_{\boldsymbol{p}_{\mathrm{obs}} \in \mathcal{P}_f} \|\boldsymbol{p}_t - \boldsymbol{p}_{\mathrm{obs}}\|\), where $\mathcal{P}_f$ is the filtered point cloud containing only the closest LiDAR point in each Cell $(i,j)$. The instantaneous collision penalty is then computed as:
\begin{equation}
\begin{aligned}
J_{\mathrm{col},t} &=
\begin{cases}
C, & d_t < d_{\mathrm{obs}}^{\min}, \\[4pt]
C\,\exp\!\big(-a(d_t - d_{\mathrm{obs}}^{\min})\big), & d_{\mathrm{obs}}^{\min} \le d_t < d_{\mathrm{obs}}^{\max}, \\[4pt]
0, & d_t \ge d_{\mathrm{obs}}^{\max},
\end{cases} \\[8pt]
J_{\mathrm{col}} &= \sum_{t=0}^{N-1}J_{\mathrm{col},t},
\end{aligned}
\end{equation}
where $d_{\mathrm{obs}}^{\min}$ and $d_{\mathrm{obs}}^{\max}$ denote the minimum and maximum safe distances, satisfying $d_{\mathrm{obs}}^{\min} < d_{\mathrm{obs}}^{\max}$; $C>0$ is a scaling factor, and $a>0$ is the slope parameter.

\section{Experiment}

\subsection{Implementation Choice}
We validate AERO-MPPI extensively in both simulation and real-world environments. For practical deployment, the controller must run at high frequency with efficient rollout computation. However, CPU execution of MPPI is prohibitive due to the overhead of point-cloud nearest-neighbor costs. A pure CUDA implementation~\cite{vlahov2024mppi} can deliver high efficiency but is difficult to prototype and maintain because it demands low-level GPU programming, manual memory management, and is tied to NVIDIA hardware. In contrast, JAX~\cite{jax2018github} provides automatic differentiation and a more productive Python environment but lacks low-level GPU control. We therefore adopt NVIDIA Warp~\cite{warp2022}, which combines near-CUDA efficiency with concise Python APIs, PyTorch integration, and explicit kernel control. For implementation, all modules of AERO-MPPI are written as Warp kernels to fully exploit GPU parallelism and ensure real-time performance. LiDAR point clouds are accumulated over the most recent 10 consecutive frames, and \(M = 15\   \bigl(M_h = 5,\; M_v = 3,\; \text{corresponding to searching a field of view } 90^\circ \times 54^\circ \bigr)\)
anchor points are sampled, seeding \(15\) parallel MPPI instances. The other parameters are summarized in Table~\ref{tab:mppi_params}.

\begin{table}[h]
\caption{MPPI and AERO-MPPI Parameters}
\label{tab:mppi_params}
\centering
\begin{tabular}{lc}
\toprule
\multicolumn{2}{c}{\textbf{MPPI Parameters}} \\
\midrule
$m$                & $1.0\,\text{kg}$ \\
$F_{t,min}$         & $0.3\,\text{N}$ \\
$F_{t,max}$         & $16.35\,\text{N}$ \\
$\omega_{xy,max}$  & $3.0\,\text{rad/s}$ \\
$\omega_{z,max}$   & $2.0\,\text{rad/s}$ \\
$\Sigma_{{F_t}}$ & $1.0$ \\
$\Sigma_{\omega_{xy}}$  & $1.0$ \\
$\Sigma_{\omega_{z}}$   & $0.5$ \\
$\Delta t$         & $50\,\text{ms}$ \\
$K$                & $128$ \\
$N$                & $25$ \\
$\lambda$          & $0.1$ \\
\bottomrule
\end{tabular}%
\hspace{2em}
\begin{tabular}{lc}
\toprule
\multicolumn{2}{c}{\textbf{AERO-MPPI Parameters}} \\
\midrule
$Q_{p}$                 & $3.0$ \\
$Q_{v}$                 & $0.25$ \\
$Q_{q}$                 & $1.0$ \\
$Q_{\text{vnorm}}$      & $0.15$ \\
$Q_{\text{track}}$      & $15.0$ \\
$Q_c$                     & $0.5$ \\
$Q_{c\Delta}$            & $0.5$ \\
$\ell$                  & $5.0\,\text{m}$ \\
$C$                     & $10^{6}$ \\
$a$                & $5.0$ \\
$d_{obs}^{min}$         & $0.4\,\text{m}$ \\
$d_{obs}^{max}$         & $1.0\,\text{m}$ \\
\bottomrule
\end{tabular}
\end{table}

\begin{table*}[t]
\begin{threeparttable}
\caption{Performance comparison in three scenarios}
\label{tab:perf_comparison}
\centering
\begin{tabularx}{\textwidth}{l l YY Y Y YY}
\toprule
\multirow{2}{*}{Method} & \multirow{2}{*}{Scene} 
& \multicolumn{2}{c}{Vel. (m/s) $\uparrow$} 
& \multirow{2}{*}{Smooth (m$^2$/s$^5$) $\downarrow$}
& \multirow{2}{*}{Length (m) $\downarrow$}
& \multicolumn{2}{c}{Clearance (m)$^\ast$} \\ 
\cmidrule(lr){3-4} \cmidrule(lr){7-8}
& & Avg & Max & & & Avg & Min \\
\midrule
\multirow{3}{*}{Ego-planner~\cite{egoplanner2021}} 
& Forest   & $1.82 \pm 0.11$ & $6.37 \pm 0.17$ & $278.08 \pm 44.99$ & $52.95 \pm 2.33$ & $1.66 \pm 0.09$ & $0.43 \pm 0.14$ \\
& Verticals & $1.33 \pm 0.07$ & $3.24 \pm 0.05$ & $39.02 \pm 32.33$  & $52.60 \pm 2.13$ & $1.26 \pm 0.08$ & $0.35 \pm 0.10$ \\
& Inclines & $1.04 \pm 0.06$ & $2.31 \pm 0.72$ & $70.02 \pm 73.54$  & $57.88 \pm 3.42$ & $1.09 \pm 0.07$ & $0.31 \pm 0.05$ \\
\midrule
\multirow{3}{*}{Topo-planner~\cite{topoplanner2020}} 
& Forest   & $3.27 \pm 0.24$ & $6.64 \pm 1.08$ & $87.26 \pm 25.60$ & $46.46 \pm 0.34$ & $1.96 \pm 0.07$ & $0.32 \pm 0.15$ \\
& Verticals & $1.42 \pm 0.08$ & $3.02 \pm 0.24$ & $49.96 \pm 24.54$ & $57.90 \pm 4.37$ & $1.18 \pm 0.05$ & $0.48 \pm 0.02$ \\
& Inclines & $1.02 \pm 0.08$ & $2.70 \pm 0.84$ & $28.51 \pm 9.49$  & $59.24 \pm 4.75$ & $1.24 \pm 0.04$ & $0.49 \pm 0.04$ \\
\midrule
\multirow{3}{*}{Fast-planner~\cite{fastplanner2019robust}} 
& Forest   & $2.53 \pm 0.25$ & $7.26 \pm 0.88$ & $67.26 \pm 23.59$ & $47.05 \pm 0.93$ & $1.76 \pm 0.03$ & $0.65 \pm 0.17$ \\
& Verticals & $1.21 \pm 0.09$ & $3.13 \pm 0.57$ & $105.72 \pm 10.17$ & $59.33 \pm 3.74$ & $1.02 \pm 0.05$ & $0.37 \pm 0.10$ \\
& Inclines & $1.16 \pm 0.10$ & $2.32 \pm 0.15$ & $\mathbf{18.51 \pm 8.02}$ & $68.75 \pm 11.13$ & $1.07 \pm 0.07$ & $0.36 \pm 0.13$ \\
\midrule
\multirow{3}{*}{IPC~\cite{liu2024integrated}} 
& Forest   & $1.84 \pm 0.14$ & $3.16 \pm 0.02$ & $79.38 \pm 27.06$ & $45.72 \pm 0.64$ & $1.76 \pm 0.07$ & $0.65 \pm 0.07$ \\
& Verticals & $1.29 \pm 0.06$ & $3.23 \pm 1.31$ & $281.08 \pm 122.35$ & $62.86 \pm 6.77$ & $0.93 \pm 0.06$ & $0.33 \pm 0.08$ \\
& Inclines & $1.33 \pm 0.03$ & $3.60 \pm 0.37$ & $303.30 \pm 86.76$ & $66.09 \pm 5.71$ & $0.97 \pm 0.08$ & $0.32 \pm 0.02$ \\
\midrule
\multirow{3}{*}{Proposed} 
& Forest   & $\mathbf{5.60 \pm 0.11}$ & $\mathbf{7.74 \pm 0.22}$ & $\mathbf{59.38 \pm 27.06}$ & $\mathbf{45.92 \pm 0.55}$ & $1.27 \pm 0.05$ & $0.51 \pm 0.06$ \\
& Verticals & $\mathbf{3.52 \pm 0.21}$ & $\mathbf{6.33 \pm 0.57}$ & $\mathbf{38.20 \pm 18.55}$ & $\mathbf{51.69 \pm 2.33}$ & $0.89 \pm 0.07$ & $0.39 \pm 0.02$ \\
& Inclines & $\mathbf{5.08 \pm 0.19}$ & $\mathbf{8.09 \pm 0.66}$ & $65.30 \pm 30.76$ & $\mathbf{48.66 \pm 1.03}$ & $0.96 \pm 0.09$ & $0.33 \pm 0.02$ \\
\bottomrule
\end{tabularx}
\begin{tablenotes}
    \footnotesize
    \item[$^\ast$] Clearance (m) denotes the distance to the nearest obstacle. 
    Extremely small values indicate collision risk, while excessively large values 
    suggest conservative detours. Moderate values typically reflect a balance 
    between safety and efficiency.
  \end{tablenotes}
\end{threeparttable}
\vspace{-5mm}
\end{table*}

\subsection{Simulation Experiments}
For a fair comparison, all simulation environments initialize the drone at $(0,0,2)$ with a target at $(45,0,2)$. The LiDAR sensing parameters of all baselines are set identical to those of our framework. In mapping-based methods, the map inflation size is matched to AERO-MPPI’s minimum safety distance $d_{\mathrm{obs}}^{\min}$ for fair comparison, and all are set equal to the UAV radius.

\subsubsection{Performance across Varying Obstacle Densities}
We first evaluate the proposed AERO-MPPI in the IsaacLab simulator~\cite{isaaclab2023orbit} under environments with varying obstacle densities and heights to examine its agility and effectiveness in mitigating local minima issues. The drone navigates through environments with progressively increasing levels of clutter. 
The results are presented in Fig.~\ref{fig:different_obstacles}, which report the maximum and average flight velocities under different clutter levels. 
Notably, even in the densest and most challenging scenarios with 1,000 obstacles, AERO-MPPI consistently maintains agile forward motion and avoids being trapped in local minima or stalling in place. Beyond navigation performance, our GPU-accelerated implementation achieves real-time execution at \textbf{500\,Hz} on a GeForce RTX 4080 SUPER GPU, with moderate GPU memory usage (less than 600\,MB when $N{=}25$ and $K{=}256$), scaling linearly with the number of trajectories and horizon length.

\begin{figure}[t]
    \centering
    \includegraphics[width=6.5cm]{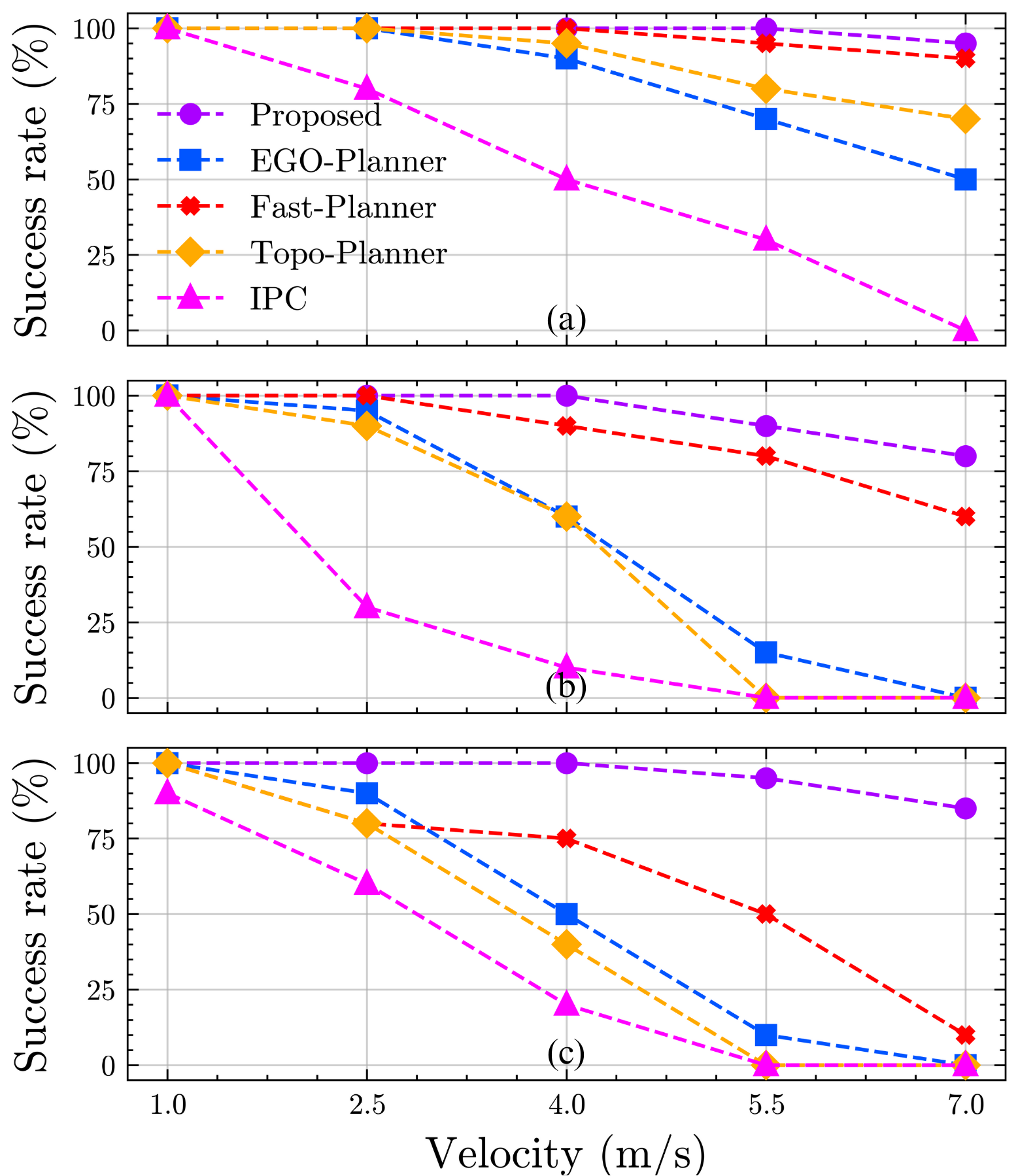}
    \vspace{-2 mm}
    \caption{
    \textbf{Benchmark results with baseline SOTA works \cite{egoplanner2021,topoplanner2020,liu2024integrated,fastplanner2019robust} under different velocity limits (e.g., 1.0–7.0 m/s) in a \(40\,\mathrm{m} \times 40\,\mathrm{m}\) map.} 
    Figures (a)–(c) illustrate the comparison results of success rate in
    scenarios Forest, Verticals, and Inclines, respectively.
    }
    \label{fig:success_rate}
    \vspace{-6 mm}
\end{figure}

\subsubsection{Comparisons with SOTA Planners under Different Cluttered Environments}
To further demonstrate AERO-MPPI’s superior performance, we benchmark it against four SOTA obstacle-avoidance systems, namely EGO-Planner~\cite{egoplanner2021}, Topo-Planner~\cite{topoplanner2020}, Fast-Planner~\cite{fastplanner2019robust}, and IPC~\cite{liu2024integrated}, in the LiDAR-based drone simulator MARSIM~\cite{marsim2023} across a variety of cluttered environments, as illustrated in Fig.~\ref{fig:simulated_comparison}.

Success rates are averaged over 50 trials (accounting for randomness from LiDAR measurement noise, algorithm initializations, etc.) for each velocity limit, as shown in Fig.~\ref{fig:success_rate}. The evaluation map size is \(40\,\mathrm{m} \times 40\,\mathrm{m}\), with three cluttered scenarios: 
\textbf{Forest}: 100 randomly scaled and tilted trees;  
\textbf{Verticals}: 1000 vertical cylinders with a height of 6\,m and radii ranging from 0.4\,m to 1.1\,m;  
\textbf{Inclines}: 800 tilted cylinders with a height of 10\,m, tilt angles ranging from $0^\circ$–$30^\circ$, and radii ranging 0.06\,m-0.3\,m.  

At low speeds, all methods achieve high success rates. However, performance degrades as velocity increases. In relatively open settings such as Forest, most baselines maintain above 50\% success at 7\,m/s. An exception is IPC~\cite{liu2024integrated}: its use of safe-flight corridor as hard MPC constraints makes the solver brittle when corridor boundaries vary rapidly, causing failures and collisions. In denser Verticals and Inclines, the limited perceptual horizon yields partial local maps. Shortest-path plans derived from these myopic maps are suboptimal; once occlusions clear, frequent replanning is triggered. This adds tracking burden and induces oscillatory flight (see Fig.~\ref{fig:simulated_comparison}). AERO-MPPI searches broader spatial area and maintains high success rates above 80\% across all scenarios, even at high speeds.

Table~\ref{tab:perf_comparison} summarizes the performance from 50 successful runs of all methods. Note that the baseline methods had to reduce their maximum speed limits to achieve 50 successful runs. Specifically, AERO-MPPI attains the highest average and peak velocities, sustaining $>5$\,m/s on average and exceeding $8$\,m/s in Inclines, while maintaining safe clearance (typically $\geq 0.3$\,m). Among baselines, EGO-Planner~\cite{egoplanner2021} shows limited agility (average speed $<2$\,m/s) and notably poor smoothness in Forest ($>270$\,m$^2$/s$^5$). Topo-Planner~\cite{topoplanner2020} achieves the largest obstacle margins (average up to $\sim1.9$\,m) but at the cost of lower speed ($1$--$3$\,m/s) and longer paths. Fast-Planner~\cite{fastplanner2019robust} reaches competitive peak speed in Forest ($>7$\,m/s) yet suffers from poor smoothness and unstable trajectories. IPC~\cite{liu2024integrated} can yield high minimum clearance ($\approx0.65$\,m), but is constrained to low maximum speeds ($<3.6$\,m/s) and long detours. 

Overall, the results demonstrate that AERO-MPPI achieves a balanced trade-off, sustaining higher speed, sufficient clearance, and smoother, shorter trajectories, thereby ensuring both safety and dynamic feasibility while delivering superior agility in cluttered environments.

\subsection{Real-world Experiment}
\subsubsection{Hardware details}
We deployed our method on a 380 mm quadrotor platform weighing 1.65 kg, equipped with a Livox Mid-360 LiDAR, a PX4-based embedded autopilot, and a Jetson Orin NX 16 GB onboard computer (see Fig.~\ref{fig:real_experiment}).
The PX4 Extended Kalman Filter (EKF) runs at 100 Hz, fusing IMU measurements with Fast-LIO2 localization~\cite{fastlio22022} to provide high-frequency pose estimates.
Using these estimates, AERO-MPPI replans control sequences at 10 Hz on the Jetson GPU, while the PX4 autopilot executes the commands at 50 Hz in a receding-horizon manner.

\subsubsection{Experiment}
\begin{figure*}[t] 
    \centering
    \includegraphics[width=1.0\textwidth]{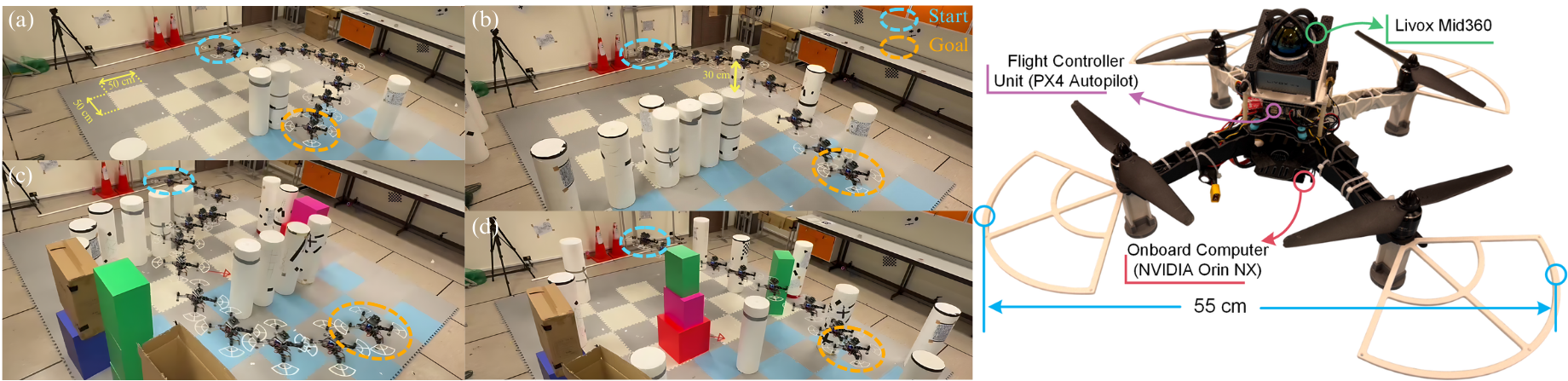}
    \vspace{-7mm}
    \caption{   
    \textbf{Indoor flight experiments in four representative scenarios.} 
    Left: four representative scenarios — (a) Pillars: navigating around multiple vertical obstacles; 
    (b) Wall: climbing over a barrier; 
    (c) Double-Wall Trap: overcoming trap configurations; 
    (d) Random Clutter: safe navigation through arbitrary obstacles. 
    Right: the hardware platform used in the experiments. 
    All experiments is initialized at 0.8\,m height and tasked to reach a goal 3.75\,m ahead.
    }
    \label{fig:real_experiment}
    \vspace{-5mm}
\end{figure*}

To further validate the real-world performance of our proposed framework, we construct four representative obstacle scenarios, and the flight results are captured in Fig.~\ref{fig:real_experiment}. The quadrotor is initially commanded to fly at a target height of 0.8 m toward a goal located 3.75 m ahead.
In the (a) \textbf{Pillar} scenario, AERO-MPPI successfully generates collision-free trajectories that steer the quadrotor around the obstacle.  
In the (b) \textbf{Wall} scenario, the framework leverages vertical trajectory samples to enable the quadrotor to climb over the wall and reach the target.  
In the (c) \textbf{Double-Wall Trap} scenario, where the obstacle layout restricts the flight altitude below the barrier and easily induces local minima, AERO-MPPI still identifies a feasible trajectory and reaches the goal.  
In the (d) \textbf{Random Clutter} scenario, with obstacles placed arbitrarily and varying inter-obstacle distances, AERO-MPPI achieves safe navigation by avoiding collisions and reaching the target reliably and smoothly. Additional hardware experiments are presented in the accompanying video. 

\section{Conclusions}
We introduced AERO-MPPI, an anchor-guided MPPI ensemble trajectory optimization framework that departs from conventional cascaded planning pipelines by unifying perception and planning into a single GPU-accelerated system. Unlike prior MPPI variants that rely on explicit maps or single-sample exploration, AERO-MPPI leverages multi-resolution spherical partitioning and anchor-guided trajectory generation to diversify exploration and mitigate local minima. This design enables the planner to discover homotopically distinct solutions that are difficult for conventional methods to capture. 
Through extensive simulation studies, we showed that AERO-MPPI sustains agile flight above 7\,m/s with success rates exceeding 80\%, consistently maintaining larger smoother trajectories compared to SOTA baselines. Real-world experiments on a LiDAR-equipped quadrotor confirmed that the entire pipeline can run fully onboard on an NVIDIA Jetson Orin NX 16\,GB, achieving reliable obstacle avoidance and robust navigation even in highly cluttered and unstructured environments.  

Overall, these results demonstrate that AERO-MPPI provides a practical and scalable solution for mapless flight in complex 3D environments. Moving forward, we plan to extend the framework toward multi-robot coordination, outdoor large-scale deployments, and integration with learning-based perception and adaptive dynamics models, aiming to further enhance robustness and autonomy in dynamic real-world scenarios.


\bibliographystyle{IEEEtran}
\bibliography{references}

\end{document}